\documentclass{article}

\usepackage{PRIMEarxiv}

\usepackage[utf8]{inputenc} 
\usepackage[T1]{fontenc}    
\usepackage{hyperref}       
\usepackage{url}            
\usepackage{booktabs}       
\usepackage{amsfonts}       
\usepackage{nicefrac}       
\usepackage{authblk}

\usepackage{microtype}      
\usepackage{lipsum}
\usepackage{fancyhdr}       
\usepackage{graphicx}       
\graphicspath{{media/}}     
\usepackage{amssymb}
\usepackage{amsmath}
\usepackage{tcolorbox}
\usepackage{xcolor}
\usepackage{subcaption}
\usepackage{graphicx, amsmath} 
\usepackage{float}
\usepackage{tikz}
\usepackage{multirow}
\usepackage{booktabs} 
\usepackage[table]{xcolor} 
\usepackage{listings}
\usepackage{xcolor}
\usepackage[toc,page]{appendix}

\usepackage{xspace}
\title{Reasoning Distillation and Structural Alignment for Improved Code Generation}

\author[1]{Amir Jalilifard}
\author[1]{Anderson de Rezende Rocha}
\author[1]{Marcos Medeiros Raimundo}

\affil[1]{Universidade Estadual de Campinas, Campinas, Brazil\\
\texttt{jalilifard@ic.unicamp.br}, \texttt{anderson.rocha@unicamp.br}, \texttt{mrai@unicamp.br}}

\begin{document}
\maketitle

\begin{abstract}
Effective code generation with language models hinges on two critical factors: accurately \textbf{understanding the intent} of the prompt and generating code that applies \textbf{algorithmic reasoning} to produce correct solutions capable of passing diverse test cases while adhering to the syntax of the target programming language. Unlike other language tasks, code generation requires more than accurate token prediction; it demands comprehension of solution-level and structural relationships rather than merely generating the most likely tokens. very large language model (VLLM) are capable of generating detailed steps toward the correct solution of complex tasks where reasoning is crucial in solving the problem. Such reasoning capabilities may be absent in smaller language models. Therefore, in this work, we distill the reasoning capabilities of a VLLM into a smaller, more efficient model that is faster and cheaper to deploy. Our approach trains the model to emulate the reasoning and problem-solving abilities of the VLLM by learning to identify correct solution pathways and establishing a structural correspondence between problem definitions and potential solutions through a novel method of structure-aware loss optimization. This enables the model to transcend token-level generation and to deeply grasp the overarching structure of solutions for given problems. Experimental results show that our fine-tuned model, developed through a cheap and simple to implement process, significantly outperforms our baseline model in terms of pass@1, average data flow, and average syntax match metrics across the MBPP, MBPP Plus, and HumanEval benchmarks.
\end{abstract}

\keywords{Large Language Models\and Coding \and Knowledge Distillation \and Chain of Thoughts}

\section{Introduction}

Large language models (LLMs) can generate text, translate from one language to another, and carry out various auto-regressive, sequence-to-sequence, and in-filling tasks, all using a single model architecture. Tasks involving logical problems require a series of steps, one after another, until the final solution arrives. However, tasks that involve a deep understanding of the problem and complex reasoning are still challenging for LLMs. Automatic coding, the task of generating codes given a programming problem, is an example of a task where logical steps should be followed until the final correct solution is reached. Compared to general language-related tasks, coding requires several complex generation abilities such as syntax and exact keyword match, algorithmic thinking, and bottom-up or top-down problem-solving capabilities. Therefore, more sophisticated approaches have been developed for training, inference, and evaluation steps of code generator language models~\cite{bavarian2022efficient, roziere2023code, su2024roformer}. The recent efforts to enhance code generation by LLMs can be roughly summarized in four major categories: a) training models with increased parameters in order to generate a richer latent space and a larger context window; b) in-context learning; c) code generation through agents interactive feedback; and d) data and instruction optimization. Each approach has advantages and disadvantages, making them appealing in specific conditions.

\textbf{In-context learning} is a subset of prompt engineering methods that aims at constructing a bridging context between the main prompt and the expected correct answer. N-shot prompts add bridging context by elucidating the task to be done with or without examples \cite{brown2020language}. One major problem with the n-shot approach is its high sensitivity to the quality, order, and relevance of the examples \cite{yoshida2024impact}, their orders, and templates \cite{lu2021fantastically, kumar2021reordering}. These complexities become even more evident when the task to be solved has different criteria of correct ordering, relevance, and importance of the given examples. 

Therefore, researchers have explored that instead of adding context, they \textbf{iteratively generate} context with LLM reasoning. Wei et al. \cite{wei2022chain} explored the contribution of a series of intermediate reasonings, called Chain-of-Thought (CoT), and concluded that it could significantly boost the complex reasoning performance of LLMs. CoTs can improve the LLM's response due to generating a sequence of intermediate steps that facilitate LLM's reasoning. However, CoT does not explore the solution space, and for more complicated problems, it may not produce the best sequence of steps toward the correct solution. To improve this shortcoming, Wei and colleagues introduced Self-consistency with CoT (CoT-SC), an ensemble of thoughts generated, and then the final response is selected using majority vote \cite{wang2022self}. Although CoT and its variants have increased the reasoning abilities of LLMs, they suffer from a lack of self-evaluation and self-improvement while generating thoughts. Tree-of-thoughts applies a self-evaluation and reasoning correction through a depth- or breadth-first-search and explores the solution space more efficiently and achieves a significant performance improvement in non-trivial logical tasks \cite{yao2023tree}. Both ToT and CoT-SC bring a broader range of solutions that improve the reasoning abilities of LLMs. However, these methods are significantly more expensive and slower than less sophisticated thought-generation methods. Some recent studies proposed slight changes to the original CoT by adding more pseudo-codes like step-by-step reasoning or adapted CoTs, which are more code structure alike \cite{li2023structured} \cite{ma2023bridging}. Despite promising results, these methods suffer from all the few-shot prompting issues mentioned above and long prompts that may be less understandable by humans, which makes debugging and interpretability very slow, if not impossible.

Another group of recent efforts has tried to use feedback loops, self-evaluation, auto-correction, and divide-and-conquer strategies to use a group of separate specialized instances of LLMs, called agents, to solve complex reasoning tasks. ToTs, for example, can be generated based on the evaluation and solution selection of two or more agents. Agents have been widely used for code generation. They can give verbal reinforcement and feedback so that the main reasoner corrects its approach toward the solution \cite{shinn2024reflexion}. These LLM instances can also take several rules similar to real-world software engineering teams to program, document, test, and return code correction and logic improvement feedbacks \cite{huang2023agentcoder} \cite{dong2024self}. Despite phenomenal problem-solving improvements by LLM agents, they are extremely costly and unduly prolonged.

Having a structured, high-quality, and massive data set of programming tasks and their associated codes is fundamental for code generation. Jain et al. \cite{jain2023llm} investigate data quality by transforming existing programs to be more structured and readable. The proposed data-cleaning pipeline includes renaming variables, modularizing complex code, and inserting natural-language-based plans via LLM-based transformations, leading to performance gains in code generation tasks. Tsai and colleagues \cite{tsai2024code} evaluated the effect of code quality and diversity versus quantity and demonstrated that a more pruned, high-quality code could reduce the computational cost and lead to better code generation. Gong et al. address the limitations of existing code search datasets by pairing high-quality queries with multiple suitable code matches. The authors collect diverse code candidates and utilize LLMs for automated annotation, filtering, and code generation, resulting in a dataset that improves model performance on code-related tasks \cite{gong2024cosqa+}. High-quality data is essential for training competent models; however, with new technologies, new versions of programming languages, and new algorithms, the data should be up to date. In such a scenario, other LLMs' effectiveness in pruning low-quality code is unknown, and manual dataset refining is time-consuming and expensive.

\textbf{Very Large language models} have a richer latent space, enabling them to capture deeper semantic and linguistic relationships within sentences, in addition, to feature extended context windows and more nuanced attention mechanisms \cite{wei2022emergent} \cite{zhao2023survey}. As model size increases, these models often demonstrate unexpected capabilities, known as emergent abilities, including enhanced instruction following and step-by-step reasoning \cite{jiang2024survey} \cite{brown2020language} \cite{kaplan2020scaling}. Although recent studies suggest that larger models do not consistently outperform smaller ones \cite{hassid2024larger}, it has been shown that code generation performance consistently improves within model families as size increases across various benchmarks \cite{roziere2023code}. Despite their advantages, VLLMs are costly to train, maintain, and deploy for inference.



In Masked Language Models (MLM) like BERT \cite{devlin2019bertpretrainingdeepbidirectional}, in different abstraction levels represent from simple similarities between word relations and their relative position in a sentence to more complex semantic and syntactic similarities \cite{anelli2022interpretability} \cite{peters2019tune}. In a code MLM, the embeddings from the last abstraction layer might capture the general meaningful coding semantic and synthetic patterns in addition to the algorithmic structure. Magister et al. investigated the possibility of training a student model to learn how to reason \cite{magister2022teaching}, yet their study is focused on tasks other than coding. In contrast to different tasks, LLMs can struggle when reasoning to solve coding tasks. This can be even more evident for tasks with higher complexity, non-trivial logical and arithmetic operators\cite{Changshu2024Reason}. Further, their method is expensive as the training is an entire parameter prone to catastrophic forgetting.

In this paper, our main contributions are as follows:
\begin{itemize}
    \item Distilling the thought-generation and problem-solving capabilities in very large language models (VLLMs) through parameter-efficient fine-tuning enables a smaller model to learn step-by-step reasoning for accurately solving coding tasks. This means teaching the smaller model to imitate the thinking process of the larger model and to learn how it reason step by step until solving the problem. In practice, it not only enables the smaller model to reason better, but also it might refine the embedding space of the smaller model.
    \item To instill an understanding of general code structure and problem-solving pathways for programming tasks, we propose a novel loss function that extends beyond basic token generation, incorporating both code structure and general algorithmic design similarities.
\end{itemize}

The primary objective of this study is to investigate how effectively reasoning distillation enables a smaller model to replicate the implicit chain-of-thought reasoning of a larger model. Additionally, we aim to explore whether a model can be trained to map clusters of similar coding problems — along with their associated reasoning patterns, algorithms, and coding styles — within the embedding space.

\section{Method}

We first formalize the general Chain-of-Thought(CoT) framework for problem-solving, upon which our approach is constructed. We then prove the learnability of automatically generated context for token-level learning. Afterward, we show how to adapt the token-level loss with the structural characteristics of code.

\subsection{Notation}

We denote a pre-trained model by \begin{math} p_{\theta} \end{math}. Let \begin{math} X = \{ x_{1}, x_{2}, \dots, x_{n} \} \end{math} represent the sequence of tokens in the task prompt, and \begin{math} Y = \{ y_{1}, y_{2}, \dots, y_{m} \} \end{math} denote the sequence of ground truth response tokens. We define \begin{math} Z = \{ z_{1}, z_{2}, \dots, z_{i-1}, z_{i} \}\end{math} as the sequence of tokens that provide auxiliary context for the model to predict the correct answer; we refer to this auxiliary context as the \emph{bridging} context.

\subsection{Reasoning distillation}

Given an initial task prompt \begin{math} X \end{math}, when the mapping between this task and the correct answer \begin{math} Y \end{math} is non-trivial, in-context prompting methods such as Chain-of-Thought (CoT) and Tree-of-Thought (ToT) propose adding a chain of reasoning steps that bridge the gap between the task and the correct answer by providing informative context \begin{math} Z \end{math}. Such context can be generated by sampling a sequence of tokens \begin{math} z_{i} \end{math} from a pre-trained model, where
\[
Z^{(k)} \sim p_{\theta}(z_{i}^{(k)} \mid X, z_{1}^{(k)}, z_{2}^{(k)}, \dots, z_{i-1}^{(k)}),
\]
With \begin{math} k \end{math} being the number of thoughts that can be generated. For ToT and CoT-SC (Self-Consistency), we have \begin{math} k > 1 \end{math}. These thoughts can be reevaluated at the end or during the thought generation process to choose the most relevant reasoning steps for a given problem. The sequence of thoughts can be sampled from the same model \begin{math} [Z^{(k)}, Y] \sim p_{\theta}(Z^{(k)}, Y \mid X) \end{math} or using independent specialized pre-trained models, followed by a feedback loop (i.e., agents to generate and evaluate in ToT).



We approach the learnability of the sequentially generated contexts within a system of independent pre-trained models, as a general framework for context generation, that collaborate to generate a series of tokens associated with a bridging context and show that such context is learnable. The original CoT and our distillation method is a specific and simpler case of such a system.

Suppose there is a system of \begin{math} Q \end{math} pre-trained VLLMs \begin{math} M = \{ m_1, m_2, \dots, m_q \} \end{math} which are sampled sequentially to generate a bridging context for a task \begin{math} T \end{math}. Each model generates tokens auto-regressively by sampling from its probability distribution \begin{math} p_{\theta_q} \end{math}, where \begin{math} \theta_q \end{math} are the parameters of model \begin{math} m_q \end{math}.

Given an initial prompt, each model \begin{math} m_q \end{math} samples a sequence of tokens \begin{math} X_n = \{ x_{n1}, x_{n2}, \dots, x_{nL_n} \} \end{math} from a conditional probability distribution:

\begin{equation}
\label{eq:model_sampling}
X_n \sim \prod_{i=1}^{L_n} p_{\theta_q}(x_{ni} \mid T, X_{<n}, x_{n1}, x_{n2}, \dots, x_{n(i-1)}),
\end{equation}

Where \begin{math} X_{<n} = \{ X_1, X_2, \dots, X_{n-1} \} \end{math} is the sequence of tokens generated by previous models, and \begin{math} L_n \end{math} is the length of the sequence generated by model \begin{math} m_q \end{math}.

Using the chain rule, the joint probability distribution over all tokens generated by the models is:

\begin{equation}
\label{eq:joint_probability}
p(T, X_1, X_2, \dots, X_K) = p(T) \prod_{n=1}^{K} \prod_{i=1}^{L_n} p_{\theta_q}(x_{ni} \mid T, X_{<n}, x_{n1}, x_{n2}, \dots, x_{n(i-1)}).
\end{equation}

For a dataset \begin{math} D = \{ (T_i, X_{1i}, X_{2i}, \dots, X_{Ki}) \}_{i = 1}^{N} \end{math} of \begin{math} N \end{math} tasks and their generated contexts, the log-likelihood of the observed data under the joint probability distribution is given by:

\begin{equation}
\label{eq:log_likelihood}
\mathcal{L}(\theta) = \sum_{i=1}^{N} \log p(T_i, X_{1i}, X_{2i}, \dots, X_{Ki}),
\end{equation}

which can be expanded using the chain rule from Equation~\eqref{eq:joint_probability}:

\begin{equation}
\label{eq:expanded_log_likelihood}
\mathcal{L}(\theta) = \sum_{i=1}^{N} \left[ \log p(T_i) + \sum_{n=1}^{K} \sum_{j=1}^{L_n} \log p_{\theta_q}(x_{nij} \mid T_i, X_{<n,i}, x_{ni1}, x_{ni2}, \dots, x_{n i(j-1)}) \right],
\end{equation}

where \begin{math} X_{<n,i} = \{ X_{1i}, X_{2i}, \dots, X_{(n-1)i} \} \end{math}.

Our optimization objective is to maximize the log-likelihood \begin{math} \mathcal{L}(\theta) \end{math}, which is equivalent to minimizing the negative log-likelihood of the generated contexts given their associated tasks. Assuming that the true joint probability distribution is \begin{math} p \end{math} and we want to approximate it using \begin{math} p_{\theta} \end{math}, the Kullback-Leibler (KL) divergence between these two distributions is:

\begin{equation}
\label{eq:kl_divergence}
D_{KL}(p \parallel p_{\theta}) = \sum_{x \in \mathcal{X}} p(x) \log \frac{p(x)}{p_{\theta}(x)} = \sum_{x \in \mathcal{X}} p(x) [\log p(x) - \log p_{\theta}(x)],
\end{equation}

where \begin{math} \mathcal{X} = \{ T, X_1, X_2, \dots, X_K \} \end{math}.

This can be expressed as:

\begin{equation}
\label{eq:kl_divergence_expectation}
D_{KL}(p \parallel p_{\theta}) = - \mathbb{E}_{p} [ \log p_{\theta}(x) ] + H_p,
\end{equation}

where \begin{math} H_p = - \sum_{x \in \mathcal{X}} p(x) \log p(x) \end{math} is the entropy of the true distribution, and \begin{math} - \mathbb{E}_{p} [ \log p_{\theta}(x) ] \end{math} is the expected negative log-likelihood under the true distribution.

Since \begin{math} H_p \end{math} does not depend on \begin{math} \theta \end{math}, minimizing \begin{math} D_{KL}(p \parallel p_{\theta}) \end{math} is equivalent to minimizing the expected negative log-likelihood \begin{math} - \mathbb{E}_{p} [ \log p_{\theta}(x) ] \end{math}. Therefore, by maximizing the log-likelihood \begin{math} \mathcal{L}(\theta) \end{math}, we are effectively minimizing the KL divergence between the true distribution \begin{math} p \end{math} and our model distribution \begin{math} p_{\theta} \end{math}. The joint probability distribution can be approximated effectively with sufficient data and appropriate model capacity.

Therefore, the bridging context \begin{math} Z \end{math} is learnable through maximizing the log-likelihood of the joint distribution of tasks and generated contexts under three conditions: a) the transformer model which tries to learn such joint probability distribution should have enough number of parameters; b) the context size should be less than the context window of such a model; and c) there should be enough training examples. It is worth mentioning that despite a theoretical foundation, practical complexities may arise while learning contexts generated by a group of independent specialized language models.

\subsection{Learning objective}
During the parameter efficient fine-tuning, we minimize the following loss function:

\begin{equation}
\label{eq:loss}
\mathcal{L} = \alpha \times \mathcal{L}_{token} + \beta \times \mathcal{L}_{s}
\end{equation}

where \begin{math} \mathcal{L}_{token} \end{math} represents token-level loss presented in Equation \ref{eq:log_likelihood}), calculated over the CoTs generated by a VLLM, the ground truth code, and a set of test cases. The structural loss, \begin{math} \mathcal{L}_{s} \end{math}, quantifies the cosine distance between the embeddings of generated code and ground truth code:

\begin{equation}
\label{eq:structure_loss}
\mathcal{L}_{s} = 1 -  \frac{E_{gt} \cdot E_{gen}}{\left| E_{gt} \right| \times \left| E_{gen} \right|}
\end{equation}

Here, \begin{math} E_{gt} \end{math} and \begin{math} E_{gen} \end{math} denote the embeddings of ground truth and generated code, respectively, computed using CodeBERT \cite{feng2020codebertpretrainedmodelprogramming}.

The weights \textit{$\alpha$} and \textit{$\beta$} are normalized parameters adjusted dynamically based on Curriculum Learning principles \cite{bengio2009curriculum}, with \textit{$\beta = 1 - \alpha$}. Early in training, we emphasize token-level accuracy by setting a higher \textit{$\alpha$}; as training progresses, \textit{$\beta$} gradually increases, shifting focus towards structural alignment over exact token matching. To ensure both token and structure losses remain active throughout training without reducing either to zero, we set predefined lower bounds for both \textit{$\alpha$} and \textit{$\beta$}. The loss calculation remains active for the CoT, code, and test cases, while attention weights are applied across the entire prompt.


\subsection{Data}
We used the Taco dataset \cite{li2023taco}, which consists of a set of 26443 Python coding questions and test cases. The questions are gathered from a variety of programming sources such as CodeForce and Leetcode. Since each question may have several correct answers submitted by different users, each question may have more than one correct solution. We chose questions for which the total number of token for questions, codes, and test cases altogether is less than 2000 tokens. This pre-processing step resulted in a total of 15360 questions and their associated first correct solution. Since each question may have more than one correct answer, we randomly chose 3000 questions from this set and then, from the remaining solutions, chose one for each question randomly. We used these alternative solutions to familiarize the model with different ways of solving the same question. Our final dataset consists of 18360 programming tasks. These questions have three different difficulty levels including easy, medium, and hard programming questions. The questions are generally long (which is challenging for LLMs) and consist of some examples, in addition to the main task description.

\section{Experiment setup and preliminary analysis}
Studies have shown that relevant contexts such as CoTs can enhance LLMs' commonsense and arithmetic reasoning performance. However, it is unclear how human language CoTs influence code generation. 

Before further experimentation, we first evaluated how context quality impacts the quality of generated code in terms of correctness and closeness to a valid solution. To minimize the effect of context size on model responses, we randomly selected 100 questions from the MBPP's \cite{austin2021program} training dataset, which includes shorter programming questions than those found in the Taco dataset. We tasked \textbf{Llama 3.1 8B} Instruct to create contexts with all previously mentioned elements. Combined with their corresponding questions, these contexts were then provided to Code \textbf{Llama 7B} \cite{roziere2023code}, a smaller model from the same family, to generate solution code. To measure how closely the generated code matched the correct solution, we employed a \textit{voting prompt} \cite{yao2023tree} and asked \textbf{GPT-4} and \textbf{Claude 3.5} Sunnet to rate both the generated context and code on a scale from 0 to 10, where 0 represents an entirely irrelevant solution and 10 a completely correct solution. Interestingly, the Pearson correlation coefficients for \textbf{GPT-4} and Claude \textbf{3.5 Sunnet} were 0.65 and 0.60, respectively, indicating a significant correlation between the context's quality and the generated code's quality. Further, the correlation for intermediate scores (4 to 7) is 0.31 and 0.27 for \textbf{GPT-4} and \textbf{Claude 3.5 Sunnet}, respectively. 
This correlation score suggests moderate context and reasoning quality may lead to inconsistent code quality. Therefore, generating high-quality CoTs, problem definitions, and test cases is essential for better code generation. Such contexts can be used directly on the prompt or can be used to refine the embedding space of language models to generate better responses.

Therefore, we propose generating rich context, in terms of question understanding and reasoning, with huge language models, and using them to train a smaller language model so that it can learn how to follow a series of correct reasoning steps to generate a valid code given a programming question. To do so, we use parameter efficient fine tuning along with a new loss that helps the model learn both the sequence of correct code tokens to be generated and the general structure of code for group of similar problems using embedding similarity. Our approach is simple to implement and cheap to execute.

\subsection{Context generation}
We employ Llama 3.1 70B \cite{dubey2024llama} to generate a structured context for each problem, comprising: a) the main intention of the question, b) a sequence of algorithmic steps leading to the correct solution, c) relevant mathematical formulas if applicable, and d) potential edge cases. These contexts are generated based on the ground truth solutions identified for each question. To maintain objectivity, we provide Llama with the correct solution but instruct it to generate a step-by-step problem-solving sequence without explicitly revealing the correct code or offering direct coding hints. Additionally, we prompt Llama to explain how specific Python built-in features or libraries, as used in the ground truth code, contribute to cleaner and more efficient solutions.

\subsection{Fine-tuning}
Low-rank adaptation (LoRA) for Large Language Models \cite{hu2021lora} was employed with ranks of 16, 32, and 64, and corresponding $\alpha$ values of 8, 16, 32, 64, and 128, to train and establish the mapping between programming questions, the problem's main intention, step-by-step reasoning, code, and test cases. The training configuration included a $5 \times 10^{-5}$ learning rate, 20 warm-up steps, a dropout rate of 0.1, 16-bit floating-point precision, and the Adam optimizer. Each model was evaluated using various benchmarks. The best performance was achieved when the LoRA rank and $\alpha$ were set to 32, although consistent improvements were observed across all configurations. Llama 3.1 8B was used as the base model for fine-tuning.

\section{Results and discussion}
We evaluated our model across several benchmarks, including MBPP, MBPP Plus, and HumanEval \cite{chen2021evaluating}, selecting pass@1 as the primary evaluation metric. Pass@1 indicates the probability that a model correctly passes all test cases on its first attempt. Due to the known limitations associated with few-shot prompting, all evaluations were conducted under zero-shot settings. We first compared our model's performance against the base model, followed by comparisons against smaller LLMs with up to 16 billion parameters. Our results are organized into two categories: 1) Context Distillation (CD) alone, and 2) Context Distillation combined with our proposed structure-aware loss. Given that MBPP Plus is a newer variant of MBPP with limited reported results in the literature, we restrict our comparative analysis with smaller LLMs to the HumanEval and MBPP benchmarks.

\begin{table}[H]
    \centering
    \begin{tabular}{p{6cm} c c c}
        \toprule
        \textbf{Model (zero-shot)} & \textbf{MBPP+} & \textbf{MBPP} & \textbf{HumanEval} \\
        \midrule
        Llama 3.1 8B (baseline model) & 48.2 & 37.67 & 21.95 \\
        \hline
        Ours - Only context distillation & \textbf{56.31} & \textbf{42.85} & \textbf{28.83} \\
        Ours - structure-aware context distillation & \textbf{56.86} & \textbf{42.42} & \textbf{35.86} \\
        \bottomrule
    \end{tabular}
    \caption{Pass@1 performance on zero-shot setting compared to the baseline model.}
    \label{table:base_model_comparison}
\end{table}

\begin{table}[htbp]
    \centering
    \begin{tabular}{p{6.5cm} l l}
        \toprule
        \textbf{Model (zero-shot)} & \textbf{MBPP} & \textbf{HumanEval} \\
        \midrule
        CODEGEN-MONO 2.7B \cite{nijkamp2022codegen} & 27.31 & 33.4\\
        CODEGEN-MONO 6.1B \cite{nijkamp2022codegen} & 32.48 & --\\
        CodeGeeX 13B \cite{zheng2023codegeex} & 22.9 & 24.4 \\
        SantaCoder 1.1B \cite{allal2023santacoder} & 14.0 & 35.0 \\
        StarCoder 15.5B \cite{li2023starcoder} & 33.0 & 52.0 \\
        WizardCoder 15B \cite{luo2023wizardcoder} & 51.8 & 57.3 \\
        Phi1 1.3B \cite{gunasekar2023textbooks} & 55.5 & 50.6 \\
        INCODER 6B \cite{fried2022incoder} & 21.30 & --\\
        Code-Cushman-001 2.5B \cite{achiam2023gpt} & 45.90 & 33.5\\
        Code Llama 7B \cite{roziere2023code} & 25.1 & 26.1\\
        Llama 3.1 8B \cite{dubey2024llama} & 37.67 & 21.95 \\
        \hline
        Ours - only context distillation & 42.85 & 28.83\\
        Ours - structure-aware context distillation & 42.42 & 35.86 \\
        \bottomrule
    \end{tabular}
    \caption{Pass@1 performance on the zero-shot setting of small models on MBPP and HumanEval datasets.}
    \label{table:complete_comparison}
\end{table}

As shown in Table \ref{table:base_model_comparison}, our fine-tuned model demonstrates notable improvements across all benchmarks. The distillation of the previously described context contributed to consistent performance gains on each benchmark. Specifically, incorporating approximately 14\% structure-aware loss resulted in a significant performance boost on HumanEval, alongside a slight improvement on MBPP Plus, though it led to a marginally reduced pass@1 performance on MBPP. Additionally, we observed a performance increase of 9\% (pass@1 of 34\%) on MBPP and 4\% (pass@1 of 30\%) on HumanEval when using Code Llama 7B as the base model. These results suggest that our model can improves the correct code generation in its first attempt between 4\% to 14\% depending on the benchmark.

Table \ref{table:complete_comparison} further highlights the performance of our model in comparison with other small models. Unlike these alternatives, our fine-tuned model is significantly more cost-effective to train and offers competitive performance. The Phi1 \cite{gunasekar2023textbooks} model with 1.3B is the only model smaller than ours that has better performance in both MBPP and HumanEval, yet our model is significantly cheaper due to using a straightforward parameter efficient fine-tuning and without any need to construct text-book quality training datasets and training from scratch. Further, our model is based on a vastly used open-source model in dozens of studies and fine-tuning efforts. Using such a well-known open-source model brings new possibilities like using model merging methods and task vector arithmetic to create new models with multitasking abilities and more fair and unbiased results \cite{ilharco2022editing}. WizardCoder \cite{luo2023wizardcoder} uses a complex process to generate a training dataset using Evol-Instruct \cite{xu2023wizardlm} approach. With 15 billion parameters, its performance exceeds all the other models, yet it is highly expensive to evaluate the generated dataset and train compared to our cheap approach, which costs approximately \$50 for its entire dataset generation and fine-tuning pipeline. The synthetic dataset generation step is not only expensive, but it is also prone to bias due to reliance on GPT3.5 and may lack the diversity required to solve real-world problems. Finally, it is unclear how smaller versions of WizardCoder perform on MBPP and HumanEval benchmarks.

We evaluated our model's performance by comparing its average dataflow and syntax match similarity scores with the ground truth code. Dataflow match metric evaluates how closely the data flow structures of the generated code align with those of the reference code, considering the information passed and processed within each. A higher dataflow match score signifies a stronger similarity in data flow between the generated and reference code. The syntax match similarity evaluates the syntactic similarity of the generated code relative to the reference code. The syntactic match score measures the degree to which the generated code conforms to the syntactic conventions of the reference code. Between these two metrics, the dataflow match is more closely to the code correctness as it measures how the data is modified along the code lines and the logic behind the information flow. Although the syntax match is not highly correlated to code correctness, it can still capture the incomplete code generation and syntax errors which can also lead to an incorrect code generation. 

The experimental results outline that our model achieved significantly higher average dataflow match scores across all benchmarks. Incorporating structure-aware loss consistently improved the average dataflow and syntax match scores across all benchmarks, except for a slight decrease in HumanEval (see Table \ref{table:average_flow}).

\begin{table}[H]
\centering
\caption{Performance Metrics for average syntax match and dataflow match}
\label{table:average_flow}
\begin{tabular}{lp{4cm}p{2cm}p{2cm}}
\toprule
\textbf{Benchmark} & \textbf{Model} & \textbf{Average syntax match} & \textbf{Average dataflow match} \\
\midrule
\multirow{3}{*}{MBPP}
 & Llama 3.1 8B & 0.2183 & 0.3809 \\
 & Ours - only context distillation & \textbf{0.2369} & \textbf{0.4489} \\
 & Ours - structure-aware context distillation & \textbf{0.2427} & \textbf{0.4566} \\
\midrule
\multirow{3}{*}{MBPP Plus} 
 & Llama 3.1 8B & \textbf{0.2738} & \textbf{0.5512} \\
 & Ours - only context distillation & \textbf{0.3259} & \textbf{0.6441} \\
 & Ours - structure-aware context distillation & \textbf{0.3521} & \textbf{0.6503} \\
\midrule
\multirow{3}{*}{HumanEval} 
 & Llama 3.1 8B & 0.3033 & 0.3099 \\
 & Ours - only context distillation & \textbf{0.3076} & \textbf{0.3602} \\
 & Ours - structure-aware context distillation & 0.3019 & \textbf{0.3688} \\
\bottomrule
\end{tabular}
\end{table}

In addition, we carried out complementary analyses to investigate the effect of our fine-tuning on the understanding of questions. To do so, we divide the questions of the MBPP test set into two parts, one with 10\% of tokens that was used to ask our model to complete the question given these tokens. The other part was used to calculate the perplexity of our fine-tuned model against the base model. As illustrated in Figure \ref{fig:question-perplexity-distribution} (see the Appendix), with a lower mean perplexity of 19.82 compared to the 21.57 of the base model, it has a lower uncertainty about understanding the general context of the coding questions.

As shown in this example, our model has a more complete understanding of the problem to be solved, and unlike the base model, our model generates a code that considers both increasing and decreasing monotonic arrays (see the Appendix).

We also evaluated the performance of the base model when CoTs generated by the Llama 3.1 70B model are added directly as the MBPP questions' context and passed to the Llama 3.1 8B. The results of our experiment show that the base modllms-hallucinations-attentionel reaches a pass@1 of 64.7\% when it uses the CoTs of very large models. In comparison, the Llama 3.1 70B model itself attains a pass@1 of 83\%. Our analysis revealed that the Llama 3.1 8B model struggles to understand the intent of questions that feature complex phrasing and lacks the mathematical knowledge required to generate code involving mathematical formulas. While our fine-tuned model improves the general comprehension of a question’s main intent, it does not significantly enhance the model’s mathematical reasoning capabilities.

\section{Conclusion}
This work investigates the feasibility of distilling key capabilities of large language models—such as intent recognition, step-by-step reasoning, and handling edge cases—through cost-effective and parameter-efficient fine-tuning. We propose a novel loss function designed to simultaneously capture fine-grained token-level information and broader semantic and structural relationships, facilitating effective mappings between programming tasks and corresponding code solutions. Our experiments indicate consistent performance improvements over the base model across multiple metrics and benchmarks. Specifically, our analysis shows that the fine-tuned model better comprehends programming questions and can more accurately predict or complete questions given partial context compared to the base model. Additionally, we demonstrate that incorporating high-quality chains-of-thought (CoTs), generated by very large language models (VLLMs), significantly enhances small-model code generation performance. Furthermore, our results indicate that fine-tuning enables smaller models to internalize these high-quality contexts to some extent. Despite achieving notable improvements across several benchmarks, the fine-tuned small model still encounters difficulties in fully grasping complex question intent and accurately interpreting mathematical formulations required by certain MBPP tasks. Future research will focus on addressing these challenges and further exploring methods to distill richer context information from VLLMs into smaller models. We are also interested in investigating the possibility of distilling the context generated through a sequential collaboration of a set of agents based on the general framework of context distillation presented in the current work.

\bibliographystyle{plain}
\bibliography{ref}

\begin{appendices}
\label{appendix:1}
We conducted additional analyses beyond pass@1 to better understand our model's behavior relative to the Llama 3.1 8B baseline. Our model demonstrates a clearer understanding of the questions and shows reduced confusion during token generation.

\begin{figure}[H]
\includegraphics[width=12cm]{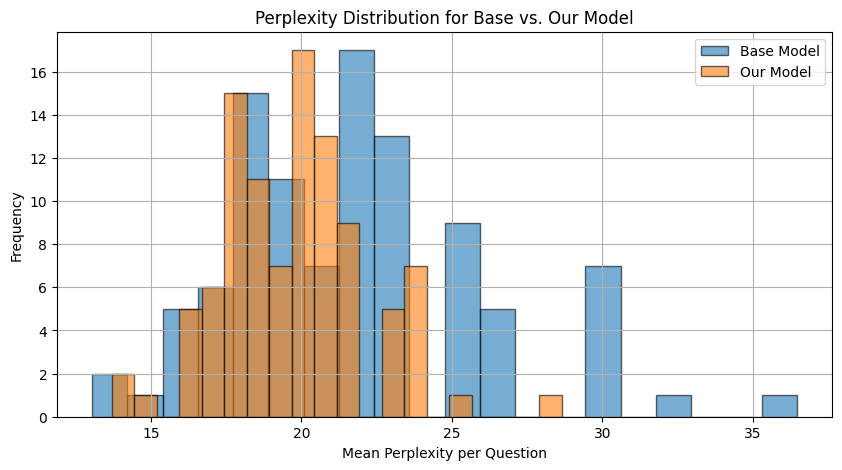}
\caption{The perplexity of our model in understanding the coding question compared to base model.}
\label{fig:question-perplexity-distribution}
\end{figure}

\begin{tcolorbox}[colback=white,colframe=black!75!black,title= An example of MBPP's question and both model's answers]

Question:
\vspace{0.5em}

\textit{Write a Python function to check whether the given array is monotonic or not.}

\vspace{0.5em}

Base model's code:
\begin{lstlisting}
def is_monotonic(arr):
    for i in range(1, len(arr)):
        if arr[i] < arr[i - 1]:
            return False
    return True
\end{lstlisting}

\vspace{0.5em}
Our model's code:

\begin{lstlisting}
def is_monotonic(arr):
    return (
        all(arr[i] <= arr[i + 1] for i in range(len(arr) - 1)) or
        all(arr[i] >= arr[i + 1] for i in range(len(arr) - 1))
    )
\end{lstlisting}

\end{tcolorbox}

\begin{figure}[H]
\includegraphics[width=12cm]{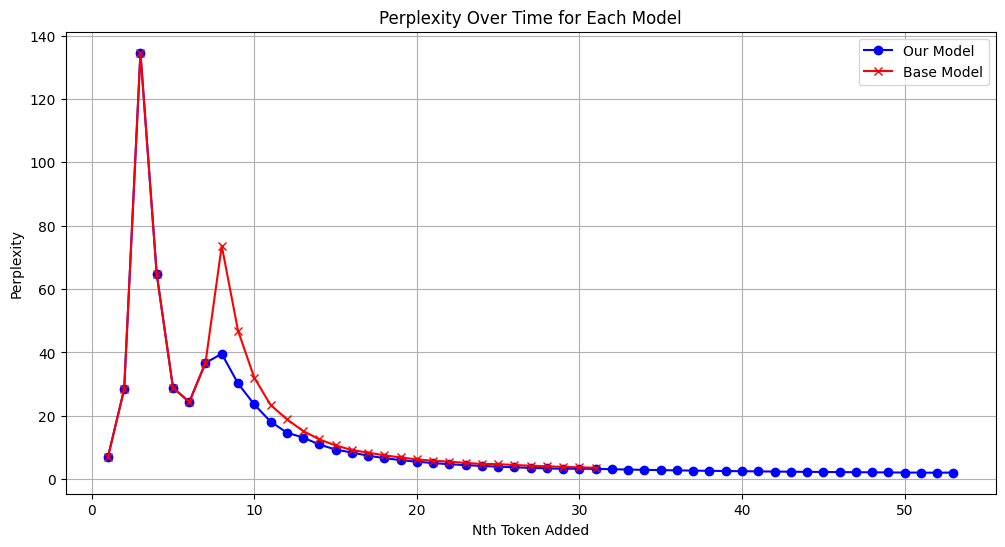}
\caption{The perplexity of our model while generating the code for a given MBPP question compared to the base model.}
\label{fig:code-perplexity}
\end{figure}

Figure \ref{fig:code-perplexity} shows the perplexity of our fine-tuned model against the base model while generating the code for one of the MBPP's questions. In general our model is less confused during the token generation. For the initial tokens both models have similar perplexities likely due to the fact that the initial tokens of a python functions are always the same when we explicitly ask the model to create a function that solve a coding problem.

\end{appendices}

\end{document}